\documentclass{article}

\usepackage{microtype}
\usepackage{graphicx}
\usepackage{subcaption}
\usepackage{booktabs} % for professional tables
\usepackage{multicol,multirow,tabularx}
\usepackage{setspace}
\usepackage{enumitem}
\usepackage{hyperref}
\usepackage{amsmath,amsfonts,bm}

\def\eqref#1{equation~\ref{#1}}
\def\1{\bm{1}}

\DeclareMathAlphabet{\mathsfit}{\encodingdefault}{\sfdefault}{m}{sl}
\SetMathAlphabet{\mathsfit}{bold}{\encodingdefault}{\sfdefault}{bx}{n}

\def\gM{{\mathcal{M}}}

\usepackage[preprint]{icml2026}

\usepackage{amsmath}
\usepackage{amssymb}
\usepackage{mathtools}
\usepackage{amsthm}
\usepackage{algorithm}

\newcommand{\ours}{A2MAML}

\usepackage[capitalize,noabbrev]{cleveref}

\theoremstyle{plain}

\theoremstyle{definition}

\theoremstyle{remark}

\usepackage[textsize=tiny]{todonotes}

\icmltitlerunning{}

\begin{document}

\twocolumn[
  \icmltitle{Active Asymmetric Multi-Agent Multimodal Learning under Uncertainty}

\icmlsetsymbol{equal}{*}

\begin{icmlauthorlist}
\icmlauthor{Rui Liu}{umd}
\icmlauthor{Pratap Tokekar}{umd}
\icmlauthor{Ming Lin}{umd}
% \icmlauthor{Firstname6 Lastname6}{sch,yyy,comp}
% \icmlauthor{Firstname7 Lastname7}{comp}
\end{icmlauthorlist}

\icmlaffiliation{umd}{University of Maryland, College Park}
% \icmlaffiliation{ncsu}{North Carolina State University}
% \icmlaffiliation{adobe}{Adobe Research}

\icmlcorrespondingauthor{Rui Liu}{ruiliu@umd.edu}

  % You may provide any keywords that you find helpful for describing your
  % paper; these are used to populate the "keywords" metadata in the PDF but
  % will not be shown in the document
  \icmlkeywords{Machine Learning, ICML}

  \vskip 0.3in
]

% this must go after the closing bracket ] following \twocolumn[ ...

% This command actually creates the footnote in the first column listing the
% affiliations and the copyright notice. The command takes one argument, which
% is text to display at the start of the footnote. The \icmlEqualContribution
% command is standard text for equal contribution. Remove it (just {}) if you
% do not need this facility.

% Use ONE of the following lines. DO NOT remove the command.
% If you have no special notice, KEEP empty braces:
\printAffiliationsAndNotice{}  % no special notice (required even if empty)
% Or, if applicable, use the standard equal contribution text:
% \printAffiliationsAndNotice{\icmlEqualContribution}

\begin{abstract}
Multi-agent systems are increasingly equipped with heterogeneous multimodal sensors, enabling richer perception but introducing modality-specific and agent-dependent uncertainty. Existing multi-agent collaboration frameworks typically reason at the agent level, assume homogeneous sensing, and handle uncertainty implicitly, limiting robustness under sensor corruption. We propose {\em Active Asymmetric Multi-Agent Multimodal Learning under Uncertainty} (A2MAML), a principled approach for uncertainty-aware, modality-level collaboration. A2MAML models each modality-specific feature as a stochastic estimate with uncertainty prediction, actively selects reliable agent–modality pairs, and aggregates information via Bayesian inverse-variance weighting. This formulation enables fine-grained, modality-level fusion, supports asymmetric modality availability, and provides a principled mechanism to suppress corrupted or noisy modalities. Extensive experiments on connected autonomous driving scenarios for collaborative accident detection demonstrate that \ours~consistently outperforms both single-agent and collaborative baselines, achieving up to \textbf{18.7\%} higher {\em accident detection} rate. 
\end{abstract}

% Our approach provides a general and practical solution for robust, uncertainty-aware multimodal multi-agent collaboration.

\section{Introduction}

The transition from single-agent systems \cite{shen2023auxiliary} to multi-agent systems \cite{liu2025caml, liu2025mmcd} represents a fundamental paradigm shift in autonomous systems. By sharing information across agents, collaborative systems can overcome limitations of single agents, such as occlusion and limited sensor range. As a result, multi-agent collaboration has become a key mechanism for improving performance and situational awareness in complex, safety-critical scenarios, such as robotics \citep{liu2025caml, arai2002advances} and autonomous driving \citep{rahman2021multi, talebpour2016influence, ye2019evaluating}.

In practice, most systems are subject to uncertainties and robustness concerns, stemming from factors such as sensor noise and environmental disturbance. Prior work in multi-agent system has explored how agents can act under uncertainty using belief updates, probabilistic reasoning, or robust policies \citep{oliehoek2016concise, foerster2018counterfactual, zhang2021multi}. However, these approaches typically assume homogeneous sensing modality shared across agents, and uncertainty is modeled at the agent or environment level rather than at the modality level. In addition, most existing multi-agent collaboration frameworks make communication and fusion decisions at the agent level, treating each agent’s sensor suite as an indivisible unit \cite{liu2020when2com, liu2020who2com, hu2022where2comm}. Such collaboration is coarse-grained, the system must either accept all modalities from a collaborator, risking contamination from corrupted sensors, or reject the agent entirely, discarding useful information from reliable modalities. Moreover, uncertainty in these communication channels is typically handled implicitly via attention weights, which lack a principled probabilistic interpretation and can lead to overconfident and poorly calibrated predictions under sensor corruption \citep{jain2019attention, guo2017calibration}. 

\begin{figure*}[t]
    \centering
    \includegraphics[width=0.9\linewidth]{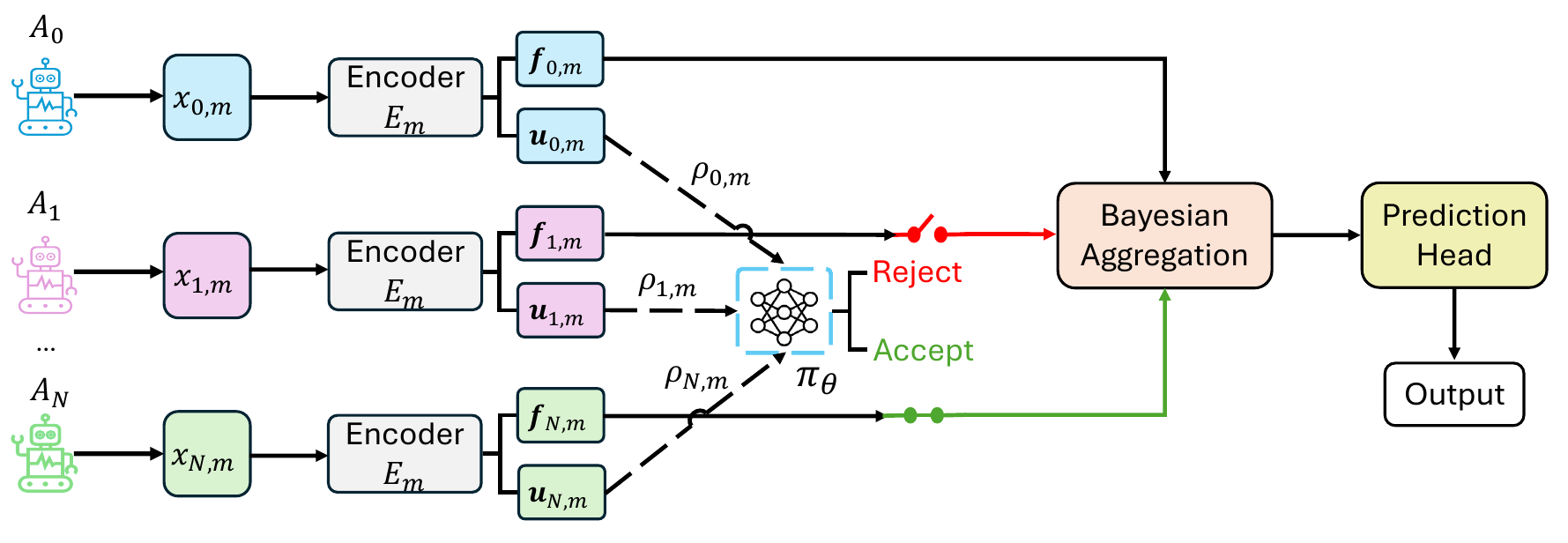}
    \vspace{-0.5em}
    \caption{\textbf{Overview of the A2MAML pipeline.} $A_0$ denotes the ego agent, while $A_1$ to $A_N$ represent collaborative agents. The framework consists of three stages. In the first stage, given corrupted raw sensor observations $\mathbf{x}_{i,m}$ from agent $i$ under modality $m$, a Gaussian feature representation is extracted, comprising a feature embedding $\mathbf{f}_{i,m}$ and an uncertainty representation $\mathbf{u}_{i,m}$. In the second stage, a scalar uncertainty token $\rho_{i,m}$ is obtained via global average pooling over $\mathbf{u}_{i,m}$, serving as a compact estimate of the sensor’s noise level. A lightweight selection policy $\pi_\theta$ takes the ego agent’s uncertainty token $\rho_{0,m}$ and a collaborative agent’s token $\rho_{i,m}$ as input to determine whether to accept or reject that modality. In the third stage, the selected feature embeddings are fused via Bayesian aggregation and passed through a prediction head to produce the ego agent’s final prediction.}
    \label{fig:app}
    % \vspace{-1em}
\end{figure*}

As autonomous systems evolve, agents are increasingly equipped with heterogeneous multimodal sensors \citep{liu2025caml, liu2025mmcd}, such as RGB cameras, LiDAR, and text. This shift fundamentally changes the nature of uncertainty in multi-agent systems. In multimodal settings, different sensing modalities exhibit distinct noise profiles and failure modes, and these effects vary across agents and over time \citep{matos2024survey}. For example, an agent’s camera may be degraded by lighting or weather while its LiDAR remains reliable, whereas another agent may experience the opposite. These varying sensing degradations introduce modality-specific and agent-dependent uncertainties, which cannot be adequately captured by existing % multi-agent
formulations.

These limitations introduce an open challenge: \textit{How can multi-agent systems collaboratively leverage heterogeneous multimodal information while explicitly modeling uncertainty, actively selecting what to communicate}? To address this problem, we propose {\em Active Asymmetric Multi-Agent Multimodal Learning under Uncertainty} (A2MAML). Unlike prior attention-based or heuristic methods, our approach supports modality-level reasoning and asymmetric modality availability. By modeling features as Gaussian distributions with learned uncertainty, we enable principled uncertainty quantification and propagation across both agents and modalities. A2MAML introduces a unified three-stage approach, as illustrated in Figure \ref{fig:app}, consisting of: (1) stochastic local encoding with modality-specific uncertainty prediction, (2) uncertainty-guided active selection, and (3) asymmetric Bayesian aggregation via inverse-variance weighting. This formulation enables fine-grained, modality-level collaboration within heterogeneous fleets and provides closed-form guarantees that unreliable modalities contribute negligibly to the fused representation. 

We validate A2MAML through extensive experiments on connected autonomous driving for collaborative accident detection. The results show that \ours~consistently outperforms both single-agent and collaborative baselines across all accident-prone scenarios, achieving up to \textbf{18.7\%} higher mean accident detection rate. These gains demonstrate the advantage of modality-level active selection and uncertainty-aware fusion, which avoid blind or agent-level aggregation and enable robust collaboration under sensor corruption. 

In summary, this work offers the following key contributions:
\begin{itemize}[left=0pt]
\vspace{-9pt}
    \item We introduce \ours, an approach that explicitly models uncertainty for modality-level and uses it to guide active selections, thereby enabling fine-grained, reliable information sharing.
    % \vspace{-6pt}
    \item We employ an uncertainty-aware asymmetric aggregation mechanism, ensuring that corrupted or noisy modalities contribute minimally to the fused representation.
    \vspace{-6pt}
    \item Through extensive experiments, we demonstrate that \ours~consistently outperforms single-agent and collaborative baselines, achieving up to {\bf 18.7\%} higher accident detection rate under challenging, sensor-corrupted scenarios.
% \vspace{-9pt}
\end{itemize}

\section{Related Work}
\subsection{Multi-Agent Collaboration}
Collaborative learning in multi-agent systems have attracted extensive attention across multiple domains. In autonomous driving, existing studies have investigated diverse coordination paradigms, ranging from LiDAR-centric end-to-end collaborative driving frameworks \citep{cui2022coopernaut} to spatio-temporal graph-based perception and control \citep{gao2024collaborative}, as well as decentralized strategies for cooperative lane changing \citep{nie2016decentralized} and decision-making grounded in game theory \citep{hang2021decision}. Collaboration has also been widely explored in robotics. For instance, \citet{mandi2024roco} introduced a hierarchical multi-robot coordination framework leveraging large language models, while \citet{zhou2022multi} proposed a graph neural network–based perception system for cooperative robots. Comprehensive surveys on collaborative robotics in search-and-rescue missions were presented by \cite{queralta2020collaborative}, and reinforcement learning approaches for multi-robot path planning have been developed in \cite{bae2019multi}. Several communication learning schemes have also been proposed, including selective communication policies such as Who2Com \citep{liu2020who2com} and When2Com \citep{liu2020when2com}, which aim to regulate what and when information should be shared.

However, existing frameworks largely treat collaboration at the agent level, forcing a binary decision to accept or reject an entire neighbor regardless of the status of their individual sensors. Furthermore, they typically assume homogeneous sensor suites across the fleet, which fails in realistic ``asymmetric'' settings where agents possess different or degraded modalities. In contrast, \ours~introduces a fine-grained, modality-level selection policy. By decoupling the decision for each sensor, our framework handles asymmetric fleets and allows the ego agent to dynamically select reliable modalities.

% \vspace{-6pt}
\subsection{Multimodal Learning}
Multimodal learning seeks to jointly model and reason over information from multiple sensory modalities to obtain richer representations. Early work explored shared latent spaces across modalities such as audio and vision \citep{ngiam2011multimodal}, while large-scale contrastive pretraining methods such as CLIP \citep{radford2021learning} enabled effective alignment between visual and textual representations. Transformer-based fusion architectures, including ViLBERT \citep{lu2019vilbert} and LXMERT \citep{tan2019lxmert}, further advanced multimodal interaction through cross-modal attention mechanisms. Beyond representation fusion, \citet{shen2023auxiliary} investigated auxiliary modality learning, where auxiliary and main modalities are jointly leveraged during training but only the main modality is used at inference, a paradigm later extended to multi-agent settings by \citet{liu2025caml}. More recently, multimodal learning has been widely explored in multimodal large language models (MLLMs) reasoning, including self-rewarding strategies \citep{li2025self}, RL based policy exploration \citep{liu2025vogue}, and compute-efficient single-rollout RL methods \citep{liu2025stable}.

Despite these advances, standard multimodal frameworks often rely on high-quality, synchronized sensor inputs. Common fusion strategies, such as concatenation or static averaging, lack the ability to identify and suppress corrupted modalities dynamically. When a sensor becomes noisy or fails (e.g., camera blackout), the corrupted signal contaminates the joint representation rather than being discarded. While robustness has been explored in certain multimodal settings, such as missing modalities \citep{liu2025caml, liu2025mmcd}, most existing methods do not leverage uncertainty for active, per-sample modality selection under corruption. Consequently, uncertainty is rarely used as a control signal to dynamically regulate collaboration and fusion in accident-prone environments. Our approach addresses this gap by introducing an active, uncertainty-driven selection mechanism that treats fusion not as a static summation, but as a dynamic, stochastic inference process.

% \vspace{-10pt}
\subsection{Learning under Uncertainty in Collaborative Systems}
Learning under uncertainty \citep{liu2023data, liu2024towards, liu2025aukt} is fundamental for autonomous agents operating in complex environments. Traditional approaches utilize Bayesian neural networks (BNNs) \cite{jospin2022hands}, monte carlo dropout \cite{milanes2021monte}, or deep ensembles to quantify uncertainty \cite{lakshminarayanan2017simple}. In collaborative systems, agents need to reason about the reliability of their peers. Works on Multi-Agent Reinforcement Learning (MARL) and planning typically model this using partially observable markov decision processes (POMDPs) or belief-space planning, where agents maintain a probabilistic belief over the joint state space to act safely under noisy observations \citep{kaelbling1998planning, bai2015intention, oliehoek2016concise}. These methods demonstrate that uncertainty quantification substantially improves robustness compared to deterministic point estimates.

However, current frameworks for learning under uncertainty exhibit limitations when applied to multimodal collaboration. First, uncertainty is typically modeled at the agent level rather than the modality level. This granularity is insufficient for asymmetric sensor failures. Second, uncertainty is often used passively, for risk-sensitive control or safety constraints, rather than actively driving the learning and fusion process. In contrast, our approach uses uncertainty as an explicit control signal to gate communication through the selection policy and to regulate multimodal fusion via inverse-variance weighting, enabling reliable learning in multi-agent multimodal systems.

% \vspace{-10pt}
\section{Approach} \label{sec:app}

\subsection{Problem Formulation} \label{sec:prob}

We first formulate the problem. Formally, let there are $N+1$ agents indexed by $i \in \{0, 1, \dots, N\}$, where $i=0$ denotes the ego agent and $i \in \{ 1, 2, \dots, N\}$ denote the collaborative agents.  The ego agent can receive information from collaborative agents within a communication range $\tau$. Let $\gM$ be the set of global modalities of all agents (e.g., $\gM=\{\text{RGB, LiDAR, Text}\}$). Each agent processes a subset of modalities $\gM_i \in \gM$, and these subsets may differ across agents. The raw observation of agent $i$ under modality $m \in \mathcal{M}_i$ is denoted as $\mathbf{x}_{i,m}$. At time $t$, the multimodal observation of agent $i$ is $\mathbf{x}_i^t = \{\mathbf{x}_{i,m}^t \mid m \in \gM_i\}$, and the joint ego--collaborator observation set is $\mathbf{X}_t = \{\mathbf{x}_0^t, \mathbf{x}_1^t, \ldots, \mathbf{x}_N^t\}$. Given $\mathbf{X}_t$, we aim to predict the ego agent’s action, such as braking for autonomous vehicles, particularly in accident-prone scenarios. 

% \vspace{-10pt}
\subsection{Active Asymmetric MAML under Uncertainty}

We model the observation process under uncertainty. We assume the raw sensor data $\mathbf{x}_{i,m}$ is a corrupted manifestation of the true environmental state $\mathbf{y}$:
\begin{equation}
    \mathbf{x}_{i,m} = f_m(\mathbf{y}) + \mathbf{\epsilon}_{i,m}, \quad  \mathbf{\epsilon}_{i,m} \sim \mathcal{N}(0, \Sigma_{i,m}(\xi)), 
\end{equation}
where the noise covariance $\Sigma_{i,m}(\xi)$ depends on the current environmental context $\xi$ (e.g., occlusion, sensor degradation, weather).

The goal of the ego agent is to estimate the posterior distribution
$p(\mathbf{y} \mid \mathbf{x}_0, \mathbf{X}_{\mathcal{S}})$,
where $\mathbf{x}_0$ denotes the ego agent’s local observation and
$\mathbf{X}_{\mathcal{S}} = \{ \mathbf{x}_{i,m} \mid Z_{i,m}=1 \}$
is the subset of neighboring observations actively selected for collaboration. We introduce a binary decision matrix $\mathbf{Z} \in \{0,1\}^{N \times |\mathcal{M}|}$, where $Z_{i,m}=1$ indicates that the ego agent requests the information of modality $m$ from collaborative agent $i$. 

% The objective is equivalent to maximization of mutual information:
% $
%     \max_{\mathbf{Z}} \ \mathcal{I}(\mathbf{y}; \mathbf{x}_0 \cup \{ \mathbf{x}_{i,m} \mid Z_{i,m}=1 \})
% $
% However, solving this directly is intractable due to the high dimensionality of $\mathbf{x}$ and the discrete nature of $\mathbf{Z}$. 

% In the following section, we propose a three-stage approach with differentiable variational relaxation.

To achieve this goal, we propose \textit{Active Asymmetric Multi-Agent Multimodal Learning under Uncertainty}, a three-stage approach with differentiable variational relaxation, consisting of: (1) stochastic local encoding, (2) uncertainty-guided active selection, and (3) asymmetric Bayesian aggregation.

% \vspace{-5pt}
\paragraph{Stochastic Local Encoding.}
To enable uncertainty-aware collaboration, agents must quantify their own reliability before communication. Standard deep networks yield deterministic point estimates, which are insufficient for identifying corruption. We employ a Gaussian encoder $\mathrm{E}_m(\cdot)$. For every input $\mathbf{x}_{i,m}$, the encoder predicts a tuple defining a Gaussian distribution over the feature space:
\begin{equation}
    \mathbf{f}_{i,m}, \mathbf{u}_{i,m} = \mathrm{E}_m(\mathbf{x}_{i,m}),
\end{equation}
where $\mathbf{f}_{i,m}$ is the feature embedding. $\mathbf{u}_{i,m}$ is the predicted uncertainty (variance) map, representing $\sigma^2_{i,m}$.

% \vspace{-5pt}
\paragraph{Uncertainty-Guided Active Selection.}
We introduce a learnable communication protocol designed to optimize the trade-off between prediction accuracy and bandwidth consumption. This process operates in two stages: a lightweight handshake to exchange metadata, followed by a granular decision-making step to request feature data.

\textbf{1) Uncertainty Compression. }
Prior to feature transmission, every agent $i$ first broadcasts a negligible ``meta-packet'' summarizing the reliability of its sensors. For each modality $m$, we compute a scalar uncertainty token $\rho_{i,m}$ by applying global average pooling to the predicted variance map $\mathbf{u}_{i,m}$.
% $
%     \rho_{i,m} = \frac{1}{H \times W} \sum_{u,v} \mathbf{u}_{i,m}^{(u,v)}.
% $
This scalar $\rho_{i,m}$ serves as a compact proxy for the overall noise level of that sensor (e.g., a high value indicates a blinded camera or occluded LiDAR).

\textbf{2) Selection Policy.}
The ego agent receives these tokens from all neighbors. Since direct computation of the mutual information I is intractable in high-dimensional feature spaces, we approximate this objective by optimizing the downstream task loss Ltask, which serves as a tractable surrogate for information gain. To decide which data is worth requesting, we employ a lightweight MLP network. The policy network $\pi_\theta$ takes the ego's own uncertainty $\rho_{0,m}$ and a neighbor's uncertainty $\rho_{i,m}$ as input pairs. It learns to estimate the utility of collaboration based on the relative uncertainty gap:
\begin{equation}
    l_{i,m} = \pi_\theta( \text{concat}[\rho_{0,m}, \rho_{i,m}] ),
\end{equation}
where $l_{i,m} \in \mathbb{R}^2$ contains the unnormalized logits for the binary actions: $k=0$ (Reject) and $k=1$ (Accept). Intuitively, the network learns to output high Accept logits when the ego is uncertain ($\rho_{0}$ is high) but the neighbor is confident ($\rho_{i}$ is low). The policy treats the communication as an unordered set rather than a fixed sequence. This ensures that the system's decisions are robust to dynamic changes in the network topology (e.g., agents entering or leaving the communication range) without requiring architecture retraining.

% We formulate the selection policy as a decoupled, pairwise scoring function. This design offers two critical advantages: \textbf{Parallel Efficiency:} Independent utility estimation allows fully vectorized computation, ensuring linear scalability and avoiding the quadratic costs of global attention. \textbf{Permutation Invariance:} The policy treats the communication as an unordered set rather than a fixed sequence. This ensures that the system's decisions are robust to dynamic changes in the network topology (e.g., agents entering or leaving the communication range) without requiring architecture retraining.

\textbf{3) Differentiable Learning of Discrete Selection.}
A standard discrete decision (Accept/Reject) is non-differentiable, which prevents backpropagation. To train this policy end-to-end, we employ a \textit{stochastic reparameterization} technique. We inject random noise $g$ into the logits to decouple the stochastic sampling process from the learnable parameters.

Let $g_0, g_1$ be i.i.d. noise samples drawn from the standard extreme value distribution. This noise ensures that the discrete samples $\text{argmax}(l+g)$ strictly follow the categorical distribution parameterized by softmax($l$) \citep{jang2016categorical}. We compute the soft selection probability $p_{i,m}$ for the Accept action using softmax:
\begin{equation}
    p_{i,m} = \frac{\exp((l_{i,m}^1 + g_1))}{\sum_{k \in \{0,1\}} \exp((l_{i,m}^k + g_k))}.
\end{equation}
This reparameterization decouples the stochasticity into the independent noise terms $g$, allowing gradients to flow efficiently through the logits $l$ during backpropagation.

For the actual transmission, we require a binary decision $Z_{i,m} \in \{0, 1\}$ rather than a probability. We first compute this discrete mask, denoted as $Z_{i,m}^{\text{hard}}$, by selecting the action with the highest noisy log-probability:
$
    Z_{i,m}^{\text{hard}} = \mathop{\mathrm{argmax}}_{k \in \{0,1\}} (l_{i,m}^k + g_k).
$
This operation results in a strict 0 or 1 value. However, the $\text{argmax}$ function is non-differentiable (it has zero gradient almost everywhere), which would normally break backpropagation. To fix this, we employ a \textit{gradient bypass} technique. We construct a surrogate variable $Z_{i,m}^{\text{train}}$ that behaves differently in the forward and backward passes:
\begin{equation}
    Z_{i,m}^{\text{train}} = \mathrm{stopgrad}(Z_{i,m}^{\text{hard}} - p_{i,m}) + p_{i,m}.
\end{equation}
When calculating predictions in the forward pass, the term $(Z^{\text{hard}} - p) + p$ mathematically cancels out to become just $Z^{\text{hard}}$. This ensures the network effectively uses a strict binary mask (0 or 1). When calculating gradients in the backward pass, the $\mathrm{stopgrad}$ operator blocks the discrete term. The gradient flows directly through the smooth probability $p_{i,m}$, allowing the optimizer to update the policy logits.

\paragraph{Asymmetric Bayesian Aggregation.}
We propose a precision-weighted fusion derived from Bayesian principles. Unlike standard attention mechanisms that infer weights from feature similarity, we treat the received feature embeddings as independent Gaussian estimates . The optimal rule for combining independent Gaussian estimates is inverse variance weighting, therefore, the aggregated feature $\mathbf{f}$ corresponding to the same modality is computed as:
\begin{equation}
    \mathbf{f} = \frac{\sum_{i=0}^N \sum_{m \in \mathcal{M}_i} Z_{i,m} \cdot \omega_{i,m} \cdot \mathbf{f}_{i,m}}{\sum_{i=0}^N \sum_{m \in \mathcal{M}_i} Z_{i,m} \cdot \omega_{i,m}}
\end{equation}
where the weighting term $\omega_{i,m}$ is the precision (inverse variance):
\begin{equation}
    \omega_{i,m} = \exp(-\mathbf{u}_{i,m})
\end{equation}
This formulation provides a double-layer of robustness: (1) Coarse-grained rejection ($Z=0$). If the policy $\pi_\theta$ deems a modality redundant, it is strictly excluded from the sum. (2) Fine-grained suppression ($\omega \approx 0$). If a modality is selected ($Z=1$) but contains local noise (e.g., a specific occluded region), the encoder predicts high uncertainty ($\mathbf{u} \to \infty$). This drives the precision $\omega \to 0$, causing the fusion layer to mathematically ignore that specific region.

We then fuse (e.g., via projection and concatenation) the aggregated feature $\mathbf{f}$ of different modalities to create a comprehensive multimodal embedding $\mathbf{F}$. We train the multi-agent, multimodal system end-to-end using the following objective:
\begin{equation}
\mathcal{L}_{\text{total}}
= \mathcal{L}_{\text{task}}(\hat{\mathbf{y}}, \mathbf{y})
+ \sum_{i=0}^{N} \sum_{m \in \mathcal{M}_i} \mathbf{u}_{i,m},
\end{equation}
where \(\hat{\mathbf{y}}\) denotes the ego agent’s final prediction obtained by passing the fused representation \(\mathbf{F}\) through a task-specific prediction head, and \(\mathcal{L}_{\text{task}}\) is the supervised task loss (e.g., cross-entropy loss) computed using the ground-truth label \(\mathbf{y}\). Crucially, gradients from \(\mathcal{L}_{\text{task}}\) backpropagate through the fusion module and the uncertainty-aware selection mechanism, thereby updating both the modality encoders and the selector policy \(\pi_\theta\). This encourages the model to assign higher weights to modalities from collaborators that improve task performance.

The second term regularizes the predicted variance \(\mathbf{u}_{i,m} \), preventing all inputs are assigned infinite uncertainty. When a modality from a collaborator provides noisy, corrupted, or misaligned features that harm the fused representation, the resulting increase in \(\mathcal{L}_{\text{task}}\) induces gradients that raise the corresponding uncertainty \(\mathbf{u}_{i,m}\), effectively down-weighting its contribution during fusion. This mechanism enables the ego agent to automatically suppress unreliable collaborators sensors.

\begin{figure}[t]
    \centering
    \includegraphics[width=\linewidth]{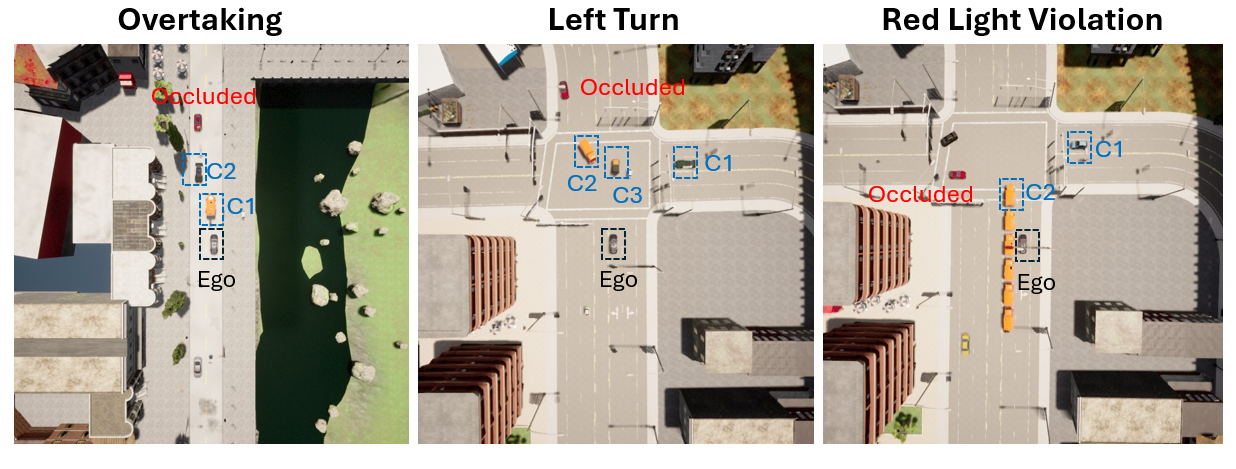}
    \caption{Accident-prone traffic scenarios in the AUTOCASTSIM benchmark for connected autonomous vehicles, including overtaking, left turn, and red light violation \citep{qiu2021autocast, liu2025caml}. Additional details for each scenario can be found in Appendix~\ref{app:data}.}
    \label{fig:scenario}
    % \vspace{-10pt}
\end{figure}

\section{Experiments}

In this section, we evaluate the effectiveness of \ours~through a series of experiments on  accident-prone connected autonomous driving scenarios. We compare it against both single-agent and collaborative baselines and assess its performance under varying levels of sensor noise. Additionally, we conduct ablation studies to quantify the contributions of key components, including modality-level active selection and Bayesian fusion.

\subsection{Data Collection}
The dataset is generated using the AUTOCASTSIM benchmark \cite{qiu2021autocast}, which comprises three realistic, complex, and accident-prone traffic scenarios designed for connected autonomous vehicles. These scenarios feature limited sensor coverage and frequent occlusions, as illustrated in Figure~\ref{fig:scenario}. Following the experimental protocol of prior work \cite{gao2024collaborative, liu2025caml}, the ego vehicle can have up to three collaborative vehicles at each timestep, provided they are within a 150-meter communication range. We collect data trails with randomized scenario configurations, where each trail includes synchronized RGB images and LiDAR point clouds from both the ego vehicle and collaborative vehicles, along with the ego vehicle’s control actions. The resulting dataset contains approximately 13K frames for training and 13K frames for testing. Additional dataset details are provided in Appendix~\ref{app:data}. 

To simulate realistic sensor corruption and communication uncertainty, we inject noise during both training and testing. Specifically, at each frame, with probability $p$, we corrupt the RGB and LiDAR observations of all vehicles. For RGB data, the corruption is randomly sampled from Gaussian noise, motion blur, or blackout. For LiDAR data, we randomly apply ray dropping or blackout.

% \vspace{-5pt}
\subsection{Experimental Setup} \label{sec:exp_setup} 
After data collection, we train all models using binary cross-entropy loss as the task loss for the ego vehicle to detect accidents at each frame, following the experimental setup of prior work \cite{gao2024collaborative, liu2025caml}.  For training, we employ a batch size of 32 and the Adam optimizer \citep{kingma2014adam} with an initial learning rate of $1e{-3}$, and a cosine annealing scheduler \citep{loshchilov2016sgdr} to adjust the learning rate over time. The model is trained on an Nvidia RTX 3090 GPU with AMD Ryzen 9 5900 CPU for 200 epochs. All experiments are repeated four times with different random seeds, and we report the mean and standard deviation of the results.

We evaluate performance using two metrics:
(1) Accident Detection Rate (ADR): the proportion of accident-prone cases correctly identified by the model among all ground-truth accident-prone cases. An accident-prone case is defined as a frame in which the expert ego vehicle executes a braking action. The objective is to assess whether the current situation is dangerous or not, enabling the ego vehicle to take braking or continue driving. (2) Expert Imitation Rate (EIR): the percentage of expert actions correctly predicted by the model across all frames, measuring how well the learned policy imitates expert driving behavior. 

For RGB observations, images are resized to $224 \times 224$ and processed using ResNet-18 \cite{he2016deep} as the stochastic local encoder, producing a 512-dimensional feature embedding along with a 512-dimensional uncertainty representation. For LiDAR observations, we adopt the Point Encoder from \citet{cui2022coopernaut} as the stochastic local encoder to obtain corresponding feature and uncertainty representations. These multimodal features are fused as described in Section~\ref{sec:app} and passed to a three-layer MLP prediction head to output the ego vehicle’s action.

\begin{figure}[t]
    \centering
    \includegraphics[width=\linewidth]{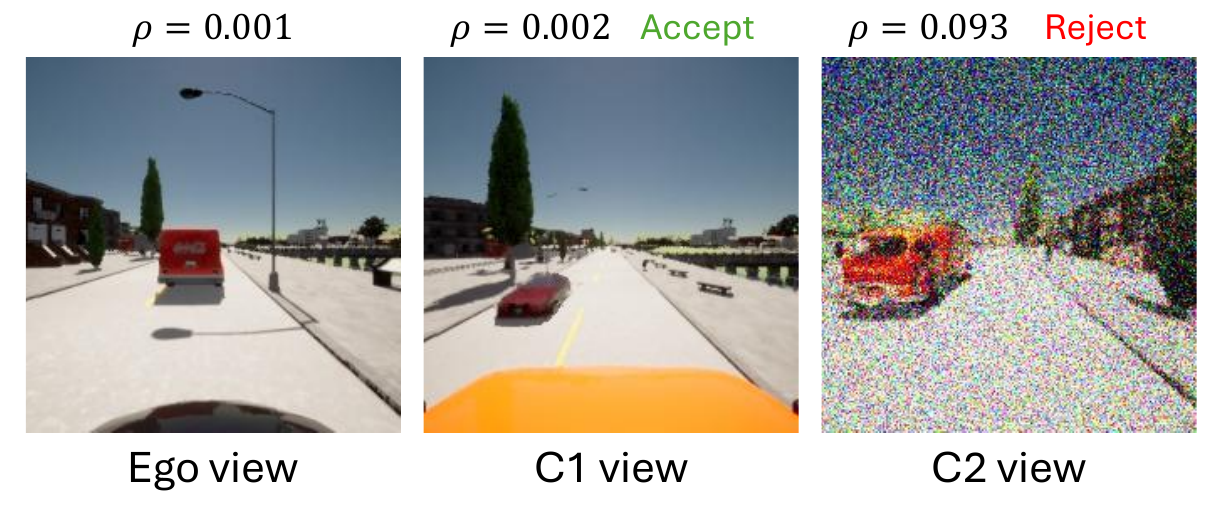}
    \caption{\textbf{Qualitative analysis of active selection.} An illustrative example demonstrates the effect of the proposed selection mechanism. The ego vehicle’s sensor data and Collaborator~1’s RGB observations are clean, while Collaborator~2’s RGB input is corrupted by Gaussian noise, resulting in a higher uncertainty token $\rho$. Consequently, the selection policy rejects the corrupted modality from Collaborator~2 while accepting the reliable features from Collaborator~1.}
    \label{fig:qual}
\end{figure}

% \vspace{-6pt}
\subsection{Baselines}
We compare A2MAML against multiple collaborative methods. To ensure fair comparison, all baselines are adapted to support multimodal inputs via intermediate fusion concatenation and replace their final detection head with the same MLP prediction head where applicable: (1) Single-Agent: The ego vehicle operates solely on its local sensors. (2) V2VNet \cite{wang2020v2vnet}: A Graph Neural Network (GNN) approach that aggregates compressed features from all neighbors, but lacks selectivity mechanism. (3) Who2com \cite{liu2020who2com}: Utilizes a handshake mechanism to learn attention weights for neighbor selection. Crucially, selection is performed at the \textit{agent level}, forcing a binary decision (accept/reject) for the entire agent regardless of partial sensor corruption. (4) When2com \cite{liu2020when2com}: Extends Who2com but relies on implicit attention rather than explicit uncertainty quantification. (5) V2X-ViT \cite{xu2022v2x}, Leverages Vision Transformers to fuse information across multiple agents.

\begin{table*}[t]
\centering
\caption{\textbf{Performance comparison (\%) of \ours~against the baseline and state-of-the-art methods.} \ours~consistently outperforms single-agent and collaborative methods across all accident-prone scenarios, achieving up to \textbf{18.7\%} higher mean ADR. Compared to V2VNet, Who2Com, and When2Com, \ours~shows superior performance by avoiding blind or agent-level fusion through modality-level active selection. It also surpasses V2X-ViT~\citep{xu2022v2x} by up to {\bf 13.1\%} in ADR and {\bf 8.0\%} in EIR, demonstrating the benefit of uncertainty-aware selective fusion under sensor corruption.}

\label{tab:res}
% \resizebox{\linewidth}{!}{
\begin{tabular}{lccccccc}
\toprule
% \multirow{2}{*}{Approach}
\multirow{2}{*}{Approach} & \multirow{2}{*}{PS (KB)} % new column
& \multicolumn{2}{c}{Overtaking}
& \multicolumn{2}{c}{Left Turn}
& \multicolumn{2}{c}{Red Light Violation} \\
\cmidrule{3-8}
& & ADR$\uparrow$ & EIR$\uparrow$
& ADR$\uparrow$ & EIR$\uparrow$
& ADR$\uparrow$ & EIR$\uparrow$ \\
\midrule

Single-Agent
& --
& 64.4$\pm$2.5 & 67.2$\pm$2.1
& 40.1$\pm$2.8 & 57.4$\pm$2.4
& 39.5$\pm$2.9 & 58.8$\pm$2.5 \\

V2VNet \citep{wang2020v2vnet}
& 4.0
& 66.8$\pm$2.2 & 69.1$\pm$2.0
& 42.5$\pm$2.4 & 58.2$\pm$2.2
& 41.2$\pm$2.6 & 61.5$\pm$2.4 \\

Who2com \citep{liu2020who2com}
& 2.5
& 67.2$\pm$1.9 & 70.5$\pm$1.8
& 44.3$\pm$2.1 & 62.1$\pm$1.9
& 42.8$\pm$2.3 & 63.2$\pm$2.1 \\

When2com \citep{liu2020when2com}
& 2.2
& 68.1$\pm$1.8 & 71.2$\pm$1.7
& 45.0$\pm$2.0 & 63.5$\pm$1.8
& 43.5$\pm$2.2 & 64.8$\pm$1.9 \\

V2X-ViT \citep{xu2022v2x}
& 4.2
& 69.5$\pm$2.0 & 72.8$\pm$1.9
& 47.1$\pm$1.7 & 64.8$\pm$2.0
& 44.1$\pm$2.3 & 66.1$\pm$1.8 \\

\ours~(Ours)
& 2.0
& 81.4$\pm$1.5 & 78.2$\pm$1.2
& 58.8$\pm$1.8 & 72.4$\pm$1.5
& 57.2$\pm$2.0 & 74.1$\pm$1.8 \\
\bottomrule
\end{tabular}
% }
\vspace*{-0.5em}
\end{table*}

\subsection{Main Results}
We corrupt the sensor data in each frame with probability 
$p=0.3$ and evaluate \ours~against the baselines. We present the results in Table \ref{tab:res}, which demonstrate a clear performance advantage of \ours~across all three accident-prone scenarios. 

Compared to the single-agent baseline, \ours~achieves substantial improvements in mean ADR across all scenarios: 17.0\% in overtaking, 18.7\% in left turn, and 17.7\% in red light violation. These gains highlight the benefits of collaborative perception, which enables the ego vehicle to aggregate complementary sensory information from connected agents. This enriched multi-agent perspective improves situational awareness, allowing the ego vehicle to detect accident-prone conditions and brake proactively to avoid collisions.

Compared to V2VNet \citep{wang2020v2vnet}, \ours~improves mean ADR by up to 16.0\%. V2VNet aggregates features from all collaborative agents without any selective mechanism; when some agents’ sensors are corrupted, blindly fusing unreliable information degrades performance.

Relative to Who2Com \citep{liu2020who2com} and When2Com \citep{liu2020when2com}, \ours~achieves up to 14.5\% and 13.8\% higher mean ADR, and 10.9\% and 9.3\% higher mean EIR, respectively. These methods perform communication selection at the agent level, enforcing a binary accept/reject decision for an entire agent regardless of modality-specific corruption, which can either discard useful sensor information or propagate corrupted signals. In contrast, \ours~operates at the modality level, actively selecting reliable sensor inputs from each agent, enabling finer-grained collaboration and more robust fusion.

Finally, compared to V2X-ViT \citep{xu2022v2x}, \ours~improves mean ADR and mean EIR by up to 13.1\% and 8.0\%, respectively, demonstrating the advantage of uncertainty-aware selective fusion over transformer-based blind aggregation under sensor corruption.

We further provide a qualitative example in Figure~\ref{fig:qual} to illustrate the effect of the active selection mechanism. While the ego vehicle’s sensor data and Collaborator~1’s RGB observations remain clean, Collaborator~2’s RGB input is corrupted by Gaussian noise, resulting in a higher uncertainty token $\rho$. Accordingly, the selection policy rejects the corrupted modality from Collaborator~2 while accepting the reliable features from Collaborator~1.

\subsection{Efficiency Analysis}
We evaluate the communication efficiency of \ours~against other collaborative methods. Following the protocol of prior work \citep{gao2024collaborative, liu2025mmcd, liu2025caml}, we adopt the shared package size (PS) metric to quantify communication bandwidth, which measures the amount of data exchanged among connected vehicles and serves as an indicator of communication cost in collaborative decision-making. The results are reported in Table~\ref{tab:res}.

As shown, \ours~achieves lower shared package size than other collaborative approaches, demonstrating superior communication efficiency. This improvement stems from its active selection mechanism, which avoids unnecessary communication by selectively collaborating with reliable agents and modalities rather than indiscriminately aggregating all available information, thereby reducing bandwidth consumption.

\subsection{Performance Analysis under Varying Noise Levels}

\begin{figure*}[t]
     \centering
     \begin{subfigure}[b]{0.28\textwidth}
         \centering
         \includegraphics[width=\textwidth]{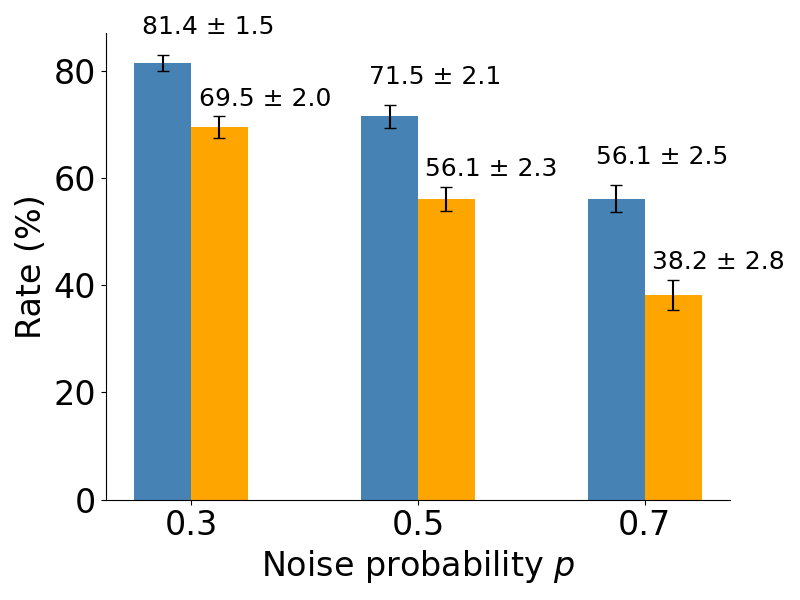}
         \caption{Overtaking}
         \label{fig:6_fle}
     \end{subfigure}
     \hfill
     \begin{subfigure}[b]{0.28\textwidth}
         \centering
         \includegraphics[width=\textwidth]{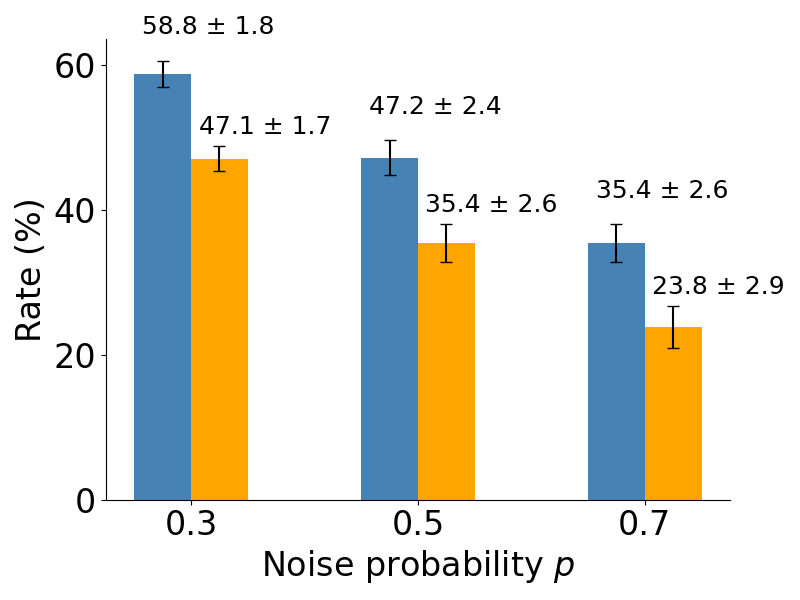}
         \caption{Left Turn}
         \label{fig:8_fle}
     \end{subfigure}
     \hfill
     \begin{subfigure}[b]{0.28\textwidth}
         \centering
         \includegraphics[width=\textwidth]{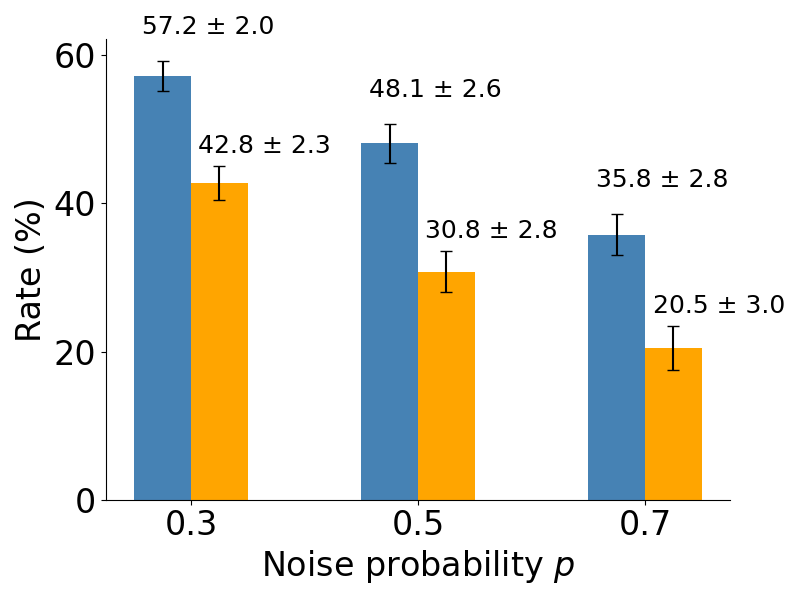}
         \caption{Red Light Violation}
         \label{fig:10_fle}
     \end{subfigure}
     \begin{subfigure}[b]{0.33\textwidth}
         \centering
         \includegraphics[width=\textwidth]{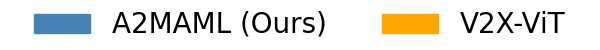}
     \end{subfigure}
     \vspace{-5pt}
     \caption{\textbf{Performance analysis on ADR under varying noise levels.} We evaluate robustness against the strong baseline V2X-ViT~\citep{xu2022v2x}, by injecting sensor noise with corruption probabilities $p=\{0.3, 0.5, 0.7\}$ during training and testing. \ours~consistently outperforms V2X-ViT across all noise levels and exhibits smaller performance degradation as noise increases, demonstrating improved robustness from modality-level active selection and uncertainty-aware Bayesian fusion.}
        \label{fig:robust}
        % \vspace{-10pt}
\end{figure*}

We conduct a performance analysis on ADR under varying noise levels to evaluate the robustness of our approach against the collaborative baseline V2X-ViT \cite{xu2022v2x}. During both training and testing, we inject noise into the sensor data of each frame with three corruption probabilities, $p=\{0.3, 0.5, 0.7\}$. The results, presented in Figure~\ref{fig:robust}, show that \ours~consistently outperforms V2X-ViT across all noise levels. While both methods experience performance degradation as noise increases, the decline for \ours~is smaller, highlighting its robustness to sensor corruption. This resilience arises from two key components of our framework: first, the modality-level active selection mechanism dynamically identifies and prioritizes reliable sensor inputs from each agent, avoiding contamination from corrupted modalities; second, the Bayesian fusion via inverse-variance weighting effectively suppresses noisy features during aggregation, ensuring that the fused representation remains accurate even under high uncertainty. These results demonstrate that explicitly modeling uncertainty and leveraging it to guide both selection and fusion significantly enhances the reliability of multi-agent multimodal decision-making in accident-prone scenarios.

% The results are presented in Figure~\ref{fig:robust}. As shown, \ours~consistently outperforms V2X-ViT across all noise levels. While the performance of both methods degrades as the noise level increases, the degradation of \ours~is smaller, demonstrating stronger robustness under sensor corruption. This resilience stems from our modality-level active selection mechanism, which identifies reliable sensor inputs, together with Bayesian fusion using inverse-variance weighting to suppress noisy features during aggregation.

\begin{table*}[ht]
\centering
\caption{\textbf{Ablation studies on active selection and Bayesian fusion.} Compared to the full method, removing active selection reduces mean ADR by up to 2.7\% and EIR by up to 3.1\%, while removing Bayesian fusion leads to a larger performance degradation of up to {\bf 9.4\%} in ADR and {\bf 7.2\%} in EIR. Eliminating both components results in the most severe decline of up to {\bf 17.3\%} in ADR and {\bf 13.5\%} in EIR, highlighting the effects of both mechanisms for robust collaborative decision-making in \ours.}
\label{tab:abl}
% \resizebox{\linewidth}{!}{
\begin{tabular}{lcccccc}
\toprule
\multirow{2}{*}{Approach}
& \multicolumn{2}{c}{Overtaking}
& \multicolumn{2}{c}{Left Turn}
& \multicolumn{2}{c}{Red Light Violation} \\
% \cmidrule(lr){2-7} \cmidrule(lr){4-5} \cmidrule(lr){6-7}
\cmidrule{2-7}
& ADR$\uparrow$ & EIR$\uparrow$
& ADR$\uparrow$ & EIR$\uparrow$
& ADR$\uparrow$ & EIR$\uparrow$ \\
\midrule

Full method
& 81.4$\pm$1.5 & 78.2$\pm$1.2
& 58.8$\pm$1.8 & 72.4$\pm$1.5
& 57.2$\pm$2.0 & 74.1$\pm$1.8 \\

% w/o Stochastic encoding
% & 85.2$\pm$1.0 & 83.2$\pm$0.9
% & 56.4$\pm$1.5 & 78.0$\pm$1.2
% & 55.8$\pm$1.3 & 81.0$\pm$1.2 \\

w/o Active selection
& 78.8$\pm$1.6 & 75.1$\pm$1.4
& 56.1$\pm$1.9 & 70.2$\pm$1.6
& 54.8$\pm$2.1 & 71.9$\pm$1.9 \\

w/o Bayesian fusion
& 72.5$\pm$1.8 & 71.0$\pm$1.5
& 49.4$\pm$2.1 & 65.5$\pm$1.8
& 48.1$\pm$2.3 & 67.2$\pm$2.0 \\

w/o Active selection \& Bayesian fusion
& 65.8$\pm$2.2 & 67.5$\pm$2.0
& 41.5$\pm$2.5 & 59.4$\pm$2.3
& 40.2$\pm$2.7 & 60.8$\pm$2.4 \\
\bottomrule
\end{tabular}
% }
\vspace*{-0.5em}
\end{table*}

\subsection{Ablation Studies}
In the ablation studies, we analyze the individual contributions of active selection and Bayesian fusion in A2MAML. We evaluate four variants: (1) the full method; (2) without active selection, where all modalities from all agents are transmitted while retaining Bayesian fusion; (3) without Bayesian fusion, where features are fused via projection and concatenation instead of inverse-variance weighting while preserving active selection; and (4) without both active selection and Bayesian fusion.

We summarize the results in Table~\ref{tab:abl}. Compared to the full method, removing active selection leads to a mean ADR drop of up to 2.7\%, while removing Bayesian fusion causes a larger degradation. Eliminating both components results in further performance decline. Overall, these results demonstrate that active modality-level selection and Bayesian uncertainty-aware fusion are complementary and jointly critical to achieving robust multi-agent multimodal collaboration.

% These findings demonstrate that both active selection and Bayesian fusion are critical to the robustness and effectiveness of \ours. 

\vspace*{-0.5em}
\section{Conclusions}

We present \ours, an active, uncertainty-aware framework for multi-agent multimodal collaboration under sensor corruption. The approach operates in three stages: stochastic local encoding, uncertainty-guided active selection, and asymmetric Bayesian aggregation. By modeling features as Gaussian distributions, \ours~captures modality-specific uncertainty, which is then used to actively select reliable sensor inputs from each agent, while Bayesian fusion via inverse-variance weighting ensures that noisy or corrupted modalities contribute minimally to the joint representation. This enables robust, fine-grained information sharing across heterogeneous agents, improving situational awareness in complex, safety-critical environments. Extensive experiments on connected autonomous driving scenarios demonstrate that \ours~consistently outperforms both single-agent and collaborative baselines, achieving up to 18.7\% higher mean accident detection rate, while maintaining robustness under varying levels of sensor noise.

% Future work may extend this framework to larger fleets, more complex multimodal sensors, and dynamic communication constraints. 

\noindent
{\bf Limitations and Future Work:  }
The characteristic of our framework is its ``double-layer'' defense mechanism, where active selection acts as a coarse-grained gatekeeper to reject uninformative neighbors, while Bayesian fusion serves as a fine-grained filter to suppress local sensory anomalies. This probabilistic formulation renders the system self-calibrating, allowing it to adapt to varying noise profiles without manual tuning.
Our current selection policy is to operate on a binary ``Accept/Reject" basis for each modality. Future work could explore adaptive compression rates, where the uncertainty token $\rho$ dictates not just if data is sent, but at what resolution it is transmitted.

\vspace*{-0.5em}
\section*{Impact Statement}

This paper presents work, whose goal is to advance the field of Multi-Agent Multimodal
Learning. There are many potential societal benefits of this work.  First, A2MAML can considerably reduce accident rates, when connected vehicles can communicate and share traffic information, such as jams, speedy cars exceeding road limits, etc.  Extending beyond vehicles, such a framework can be applicable to ensure more robust distributed smart sensing of any place and spaces, including building security, environmental monitoring, underwater observation, and more.
% none which we feel must be specifically highlighted here. \\

% In the unusual situation where you want a paper to appear in the
% references without citing it in the main text, use \nocite
% \nocite{langley00}

\bibliography{ref}

@inproceedings{gao2024collaborative,
  title={Collaborative Decision-Making Using Spatiotemporal Graphs in Connected Autonomy},
  author={Gao, Peng and Shen, Yu and Lin, Ming C},
  booktitle={2024 IEEE International Conference on Robotics and Automation (ICRA)},
  pages={4983--4989},
  year={2024},
  organization={IEEE}
}

@inproceedings{cui2022coopernaut,
  title={Coopernaut: End-to-end driving with cooperative perception for networked vehicles},
  author={Cui, Jiaxun and Qiu, Hang and Chen, Dian and Stone, Peter and Zhu, Yuke},
  booktitle={Proceedings of the IEEE/CVF Conference on Computer Vision and Pattern Recognition},
  pages={17252--17262},
  year={2022}
}

@inproceedings{shen2023auxiliary,
  title={Auxiliary modality learning with generalized curriculum distillation},
  author={Shen, Yu and Wang, Xijun and Gao, Peng and Lin, Ming},
  booktitle={International Conference on Machine Learning},
  pages={31057--31076},
  year={2023},
  organization={PMLR}
}

@article{matos2024survey,
  title={A survey on sensor failures in autonomous vehicles: Challenges and solutions},
  author={Matos, Francisco and Bernardino, Jorge and Dur{\~a}es, Jo{\~a}o and Cunha, Jo{\~a}o},
  journal={Sensors},
  volume={24},
  number={16},
  pages={5108},
  year={2024},
  publisher={MDPI}
}

@inproceedings{he2016deep,
  title={Deep residual learning for image recognition},
  author={He, Kaiming and Zhang, Xiangyu and Ren, Shaoqing and Sun, Jian},
  booktitle={Proceedings of the IEEE conference on computer vision and pattern recognition},
  pages={770--778},
  year={2016}
}

@article{qiu2021autocast,
  title={Autocast: Scalable infrastructure-less cooperative perception for distributed collaborative driving},
  author={Qiu, Hang and Huang, Pohan and Asavisanu, Namo and Liu, Xiaochen and Psounis, Konstantinos and Govindan, Ramesh},
  journal={arXiv preprint arXiv:2112.14947},
  year={2021}
}

@inproceedings{wang2020v2vnet,
  title={V2vnet: Vehicle-to-vehicle communication for joint perception and prediction},
  author={Wang, Tsun-Hsuan and Manivasagam, Sivabalan and Liang, Ming and Yang, Bin and Zeng, Wenyuan and Urtasun, Raquel},
  booktitle={European conference on computer vision},
  pages={605--621},
  year={2020},
  organization={Springer}
}

@article{hang2021decision,
  title={Decision making of connected automated vehicles at an unsignalized roundabout considering personalized driving behaviours},
  author={Hang, Peng and Huang, Chao and Hu, Zhongxu and Xing, Yang and Lv, Chen},
  journal={IEEE Transactions on Vehicular Technology},
  volume={70},
  number={5},
  pages={4051--4064},
  year={2021},
  publisher={IEEE}
}

@article{nie2016decentralized,
  title={Decentralized cooperative lane-changing decision-making for connected autonomous vehicles},
  author={Nie, Jianqiang and Zhang, Jian and Ding, Wanting and Wan, Xia and Chen, Xiaoxuan and Ran, Bin},
  journal={IEEE access},
  volume={4},
  pages={9413--9420},
  year={2016},
  publisher={IEEE}
}

@article{rahman2021multi,
  title={A multi-vehicle communication system to assess the safety and mobility of connected and automated vehicles},
  author={Rahman, Md Hasibur and Abdel-Aty, Mohamed and Wu, Yina},
  journal={Transportation research part C: emerging technologies},
  volume={124},
  pages={102887},
  year={2021},
  publisher={Elsevier}
}

@article{talebpour2016influence,
  title={Influence of connected and autonomous vehicles on traffic flow stability and throughput},
  author={Talebpour, Alireza and Mahmassani, Hani S},
  journal={Transportation research part C: emerging technologies},
  volume={71},
  pages={143--163},
  year={2016},
  publisher={Elsevier}
}

@article{ye2019evaluating,
  title={Evaluating the impact of connected and autonomous vehicles on traffic safety},
  author={Ye, Lanhang and Yamamoto, Toshiyuki},
  journal={Physica A: Statistical Mechanics and its Applications},
  volume={526},
  pages={121009},
  year={2019},
  publisher={Elsevier}
}

@article{kingma2014adam,
  title={Adam: A method for stochastic optimization},
  author={Kingma, Diederik P},
  journal={arXiv preprint arXiv:1412.6980},
  year={2014}
}

@inproceedings{liu2020when2com,
  title={When2com: Multi-agent perception via communication graph grouping},
  author={Liu, Yen-Cheng and Tian, Junjiao and Glaser, Nathaniel and Kira, Zsolt},
  booktitle={Proceedings of the IEEE/CVF Conference on computer vision and pattern recognition},
  pages={4106--4115},
  year={2020}
}

@inproceedings{liu2020who2com,
  title={Who2com: Collaborative perception via learnable handshake communication},
  author={Liu, Yen-Cheng and Tian, Junjiao and Ma, Chih-Yao and Glaser, Nathan and Kuo, Chia-Wen and Kira, Zsolt},
  booktitle={2020 IEEE International Conference on Robotics and Automation (ICRA)},
  pages={6876--6883},
  year={2020},
  organization={IEEE}
}

@article{hu2022where2comm,
  title={Where2comm: Communication-efficient collaborative perception via spatial confidence maps},
  author={Hu, Yue and Fang, Shaoheng and Lei, Zixing and Zhong, Yiqi and Chen, Siheng},
  journal={Advances in neural information processing systems},
  volume={35},
  pages={4874--4886},
  year={2022}
}

@inproceedings{mandi2024roco,
  title={Roco: Dialectic multi-robot collaboration with large language models},
  author={Mandi, Zhao and Jain, Shreeya and Song, Shuran},
  booktitle={2024 IEEE International Conference on Robotics and Automation (ICRA)},
  pages={286--299},
  year={2024},
  organization={IEEE}
}

@article{zhou2022multi,
  title={Multi-robot collaborative perception with graph neural networks},
  author={Zhou, Yang and Xiao, Jiuhong and Zhou, Yue and Loianno, Giuseppe},
  journal={IEEE Robotics and Automation Letters},
  volume={7},
  number={2},
  pages={2289--2296},
  year={2022},
  publisher={IEEE}
}

@article{queralta2020collaborative,
  title={Collaborative multi-robot search and rescue: Planning, coordination, perception, and active vision},
  author={Queralta, Jorge Pena and Taipalmaa, Jussi and Pullinen, Bilge Can and Sarker, Victor Kathan and Gia, Tuan Nguyen and Tenhunen, Hannu and Gabbouj, Moncef and Raitoharju, Jenni and Westerlund, Tomi},
  journal={Ieee Access},
  volume={8},
  pages={191617--191643},
  year={2020},
  publisher={IEEE}
}

@article{bae2019multi,
  title={Multi-robot path planning method using reinforcement learning},
  author={Bae, Hyansu and Kim, Gidong and Kim, Jonguk and Qian, Dianwei and Lee, Sukgyu},
  journal={Applied sciences},
  volume={9},
  number={15},
  pages={3057},
  year={2019},
  publisher={MDPI}
}

@article{loshchilov2016sgdr,
  title={Sgdr: Stochastic gradient descent with warm restarts},
  author={Loshchilov, Ilya and Hutter, Frank},
  journal={arXiv preprint arXiv:1608.03983},
  year={2016}
}

@article{liu2025mmcd,
  title={MMCD: Multi-Modal Collaborative Decision-Making for Connected Autonomy with Knowledge Distillation},
  author={Liu, Rui and Wang, Zikang and Gao, Peng and Shen, Yu and Tokekar, Pratap and Lin, Ming},
  journal={arXiv preprint arXiv:2509.18198},
  year={2025}
}

@article{liu2025vogue,
  title={VOGUE: Guiding Exploration with Visual Uncertainty Improves Multimodal Reasoning},
  author={Liu, Rui and Yu, Dian and Zheng, Tong and Dai, Runpeng and Li, Zongxia and Yu, Wenhao and Liang, Zhenwen and Song, Linfeng and Mi, Haitao and Tokekar, Pratap and others},
  journal={arXiv preprint arXiv:2510.01444},
  year={2025}
}

@article{li2025self,
  title={Self-rewarding vision-language model via reasoning decomposition},
  author={Li, Zongxia and Yu, Wenhao and Huang, Chengsong and Liu, Rui and Liang, Zhenwen and Liu, Fuxiao and Che, Jingxi and Yu, Dian and Boyd-Graber, Jordan and Mi, Haitao and others},
  journal={arXiv preprint arXiv:2508.19652},
  year={2025}
}

@article{liu2025aukt,
  title={AUKT: Adaptive Uncertainty-Guided Knowledge Transfer with Conformal Prediction},
  author={Liu, Rui and Gao, Peng and Shen, Yu and Lin, Ming and Tokekar, Pratap},
  journal={arXiv preprint arXiv:2502.16736},
  year={2025}
}

@inproceedings{liu2023data,
  title={Data-driven distributionally robust optimal control with state-dependent noise},
  author={Liu, Rui and Shi, Guangyao and Tokekar, Pratap},
  booktitle={2023 IEEE/RSJ International Conference on Intelligent Robots and Systems (IROS)},
  pages={9986--9991},
  year={2023},
  organization={IEEE}
}

@article{liu2024towards,
  title={Towards Efficient Risk-Sensitive Policy Gradient: An Iteration Complexity Analysis},
  author={Liu, Rui and Gupta, Anish and Noorani, Erfaun and Tokekar, Pratap},
  journal={arXiv preprint arXiv:2403.08955},
  year={2024}
}

@article{liu2025stable,
  title={Stable and Efficient Single-Rollout RL for Multimodal Reasoning},
  author={Liu, Rui and Yu, Dian and Ke, Lei and Liu, Haolin and Zhou, Yujun and Liang, Zhenwen and Mi, Haitao and Tokekar, Pratap and Yu, Dong},
  journal={arXiv preprint arXiv:2512.18215},
  year={2025}
}

@article{liu2025caml,
  title={CAML: Collaborative Auxiliary Modality Learning for Multi-Agent Systems},
  author={Liu, Rui and Shen, Yu and Gao, Peng and Tokekar, Pratap and Lin, Ming},
  journal={arXiv preprint arXiv:2502.17821},
  year={2025}
}

@inproceedings{xu2022v2x,
  title={V2x-vit: Vehicle-to-everything cooperative perception with vision transformer},
  author={Xu, Runsheng and Xiang, Hao and Tu, Zhengzhong and Xia, Xin and Yang, Ming-Hsuan and Ma, Jiaqi},
  booktitle={European conference on computer vision},
  pages={107--124},
  year={2022},
  organization={Springer}
}

@book{oliehoek2016concise,
  title={A concise introduction to decentralized POMDPs},
  author={Oliehoek, Frans A and Amato, Christopher and others},
  volume={1},
  year={2016},
  publisher={Springer}
}

@inproceedings{foerster2018counterfactual,
  title={Counterfactual multi-agent policy gradients},
  author={Foerster, Jakob and Farquhar, Gregory and Afouras, Triantafyllos and Nardelli, Nantas and Whiteson, Shimon},
  booktitle={Proceedings of the AAAI conference on artificial intelligence},
  volume={32},
  number={1},
  year={2018}
}

@article{zhang2021multi,
  title={Multi-agent reinforcement learning: A selective overview of theories and algorithms},
  author={Zhang, Kaiqing and Yang, Zhuoran and Ba{\c{s}}ar, Tamer},
  journal={Handbook of reinforcement learning and control},
  pages={321--384},
  year={2021},
  publisher={Springer}
}

@article{arai2002advances,
  title={Advances in multi-robot systems},
  author={Arai, Tamio and Pagello, Enrico and Parker, Lynne E and others},
  journal={IEEE Transactions on robotics and automation},
  volume={18},
  number={5},
  pages={655--661},
  year={2002}
}

@article{jain2019attention,
  title={Attention is not explanation},
  author={Jain, Sarthak and Wallace, Byron C},
  journal={arXiv preprint arXiv:1902.10186},
  year={2019}
}

@inproceedings{guo2017calibration,
  title={On calibration of modern neural networks},
  author={Guo, Chuan and Pleiss, Geoff and Sun, Yu and Weinberger, Kilian Q},
  booktitle={International conference on machine learning},
  pages={1321--1330},
  year={2017},
  organization={PMLR}
}

@inproceedings{ngiam2011multimodal,
  title={Multimodal deep learning.},
  author={Ngiam, Jiquan and Khosla, Aditya and Kim, Mingyu and Nam, Juhan and Lee, Honglak and Ng, Andrew Y and others},
  booktitle={ICML},
  volume={11},
  pages={689--696},
  year={2011}
}

@inproceedings{radford2021learning,
  title={Learning transferable visual models from natural language supervision},
  author={Radford, Alec and Kim, Jong Wook and Hallacy, Chris and Ramesh, Aditya and Goh, Gabriel and Agarwal, Sandhini and Sastry, Girish and Askell, Amanda and Mishkin, Pamela and Clark, Jack and others},
  booktitle={International conference on machine learning},
  pages={8748--8763},
  year={2021},
  organization={PmLR}
}

@article{lu2019vilbert,
  title={Vilbert: Pretraining task-agnostic visiolinguistic representations for vision-and-language tasks},
  author={Lu, Jiasen and Batra, Dhruv and Parikh, Devi and Lee, Stefan},
  journal={Advances in neural information processing systems},
  volume={32},
  year={2019}
}

@article{tan2019lxmert,
  title={Lxmert: Learning cross-modality encoder representations from transformers},
  author={Tan, Hao and Bansal, Mohit},
  journal={arXiv preprint arXiv:1908.07490},
  year={2019}
}

@article{jospin2022hands,
  title={Hands-on Bayesian neural networks—A tutorial for deep learning users},
  author={Jospin, Laurent Valentin and Laga, Hamid and Boussaid, Farid and Buntine, Wray and Bennamoun, Mohammed},
  journal={IEEE Computational Intelligence Magazine},
  volume={17},
  number={2},
  pages={29--48},
  year={2022},
  publisher={IEEE}
}

@article{milanes2021monte,
  title={Monte carlo dropout for uncertainty estimation and motor imagery classification},
  author={Milan{\'e}s-Hermosilla, Daily and Trujillo Codorni{\'u}, Rafael and L{\'o}pez-Baracaldo, Ren{\'e} and Sagar{\'o}-Zamora, Roberto and Delisle-Rodriguez, Denis and Villarejo-Mayor, John Jairo and Nunez-Alvarez, Jose Ricardo},
  journal={Sensors},
  volume={21},
  number={21},
  pages={7241},
  year={2021},
  publisher={MDPI}
}

@article{lakshminarayanan2017simple,
  title={Simple and scalable predictive uncertainty estimation using deep ensembles},
  author={Lakshminarayanan, Balaji and Pritzel, Alexander and Blundell, Charles},
  journal={Advances in neural information processing systems},
  volume={30},
  year={2017}
}

@article{kaelbling1998planning,
  title={Planning and acting in partially observable stochastic domains},
  author={Kaelbling, Leslie Pack and Littman, Michael L and Cassandra, Anthony R},
  journal={Artificial intelligence},
  volume={101},
  number={1-2},
  pages={99--134},
  year={1998},
  publisher={Elsevier}
}

@inproceedings{bai2015intention,
  title={Intention-aware online POMDP planning for autonomous driving in a crowd},
  author={Bai, Haoyu and Cai, Shaojun and Ye, Nan and Hsu, David and Lee, Wee Sun},
  booktitle={2015 ieee international conference on robotics and automation (icra)},
  pages={454--460},
  year={2015},
  organization={IEEE}
}

@article{jang2016categorical,
  title={Categorical reparameterization with gumbel-softmax},
  author={Jang, Eric and Gu, Shixiang and Poole, Ben},
  journal={arXiv preprint arXiv:1611.01144},
  year={2016}
}
\bibliographystyle{icml2026}

%%%%%%%%%%%%%%%%%%%%%%%%%%%%%%%%%%%%%%%%%%%%%%%%%%%%%%%%%%%%%%%%%%%%%%%%%%%%%%%
%%%%%%%%%%%%%%%%%%%%%%%%%%%%%%%%%%%%%%%%%%%%%%%%%%%%%%%%%%%%%%%%%%%%%%%%%%%%%%%
% APPENDIX
%%%%%%%%%%%%%%%%%%%%%%%%%%%%%%%%%%%%%%%%%%%%%%%%%%%%%%%%%%%%%%%%%%%%%%%%%%%%%%%
%%%%%%%%%%%%%%%%%%%%%%%%%%%%%%%%%%%%%%%%%%%%%%%%%%%%%%%%%%%%%%%%%%%%%%%%%%%%%%%
\newpage
\appendix
\onecolumn
\section{Appendix}

\subsection{Collaborative Decision Making} \label{app:cdm}
\subsubsection{Dataset Details} \label{app:data}
The dataset is generated using the AUTOCASTSIM \cite{qiu2021autocast} benchmark, which features three complex and accident-prone traffic scenarios for connected autonomous driving, characterized by limited sensor coverage or obstructed views. These scenarios are realistic and include background traffic of 30 vehicles. They involve challenging interactions such as overtaking, lane changing, and red-light violations, which inherently increase the risk of accidents: (1) \textbf{Overtaking}: A sedan is blocked by a truck on a narrow, two-way road with a dashed centerline. The truck also obscures the sedan’s view of oncoming traffic. The ego vehicle must decide when and how to safely pass the truck. (2) \textbf{Left Turn}: The ego vehicle attempts a left turn at a yield sign. Its view is partially blocked by a truck waiting in the opposite left-turn lane, reducing visibility of vehicles coming from the opposite direction. (3) \textbf{Red Light Violation}: As the ego vehicle crosses an intersection, another vehicle runs a red light. Due to nearby vehicles waiting to turn left, the ego vehicle’s sensors are unable to detect the violator. 

% Following the setup established by prior works COOPERNAUT \citep{cui2022coopernaut} and STGN \citep{gao2024collaborative}, we collect 24 data trails for each scenario, using 12 trails for training and the remaining 12 for testing.

Following prior work \cite{cui2022coopernaut, gao2024collaborative, liu2025mmcd, liu2025caml}, we employ an expert agent as configured in \citet{cui2022coopernaut}, with full access to the traffic scene, to collect data. This expert agent is deployed in the environment to generate data trails. Each trail represents a unique instantiation of the traffic scenario, with differences arising from the randomized scenario configurations. These variations include different types of vehicles present in the scene, varying initial positions and trajectories of collaborative vehicles, difference in traffic flow and interactions, resulting in different visual, LiDAR inputs and control actions for each trail. Each trail contains RGB images and LiDAR point clouds from multiple vehicles, resulting in a substantially large dataset with a total size of approximately 400 GB.

\end{document}